\pdfoutput=1

\documentclass[11pt]{article}

\usepackage{EACL2023}

\usepackage{times}
\usepackage{latexsym}

\usepackage[T1]{fontenc}

\usepackage[utf8]{inputenc}

\usepackage{microtype}

\usepackage{inconsolata}

\usepackage{booktabs}
\usepackage{multirow}

\title{MAUPQA: Massive Automatically-created \\Polish Question Answering Dataset}

\author{
    Piotr Rybak \\ \\
    Institute of Computer Science,\\
    Polish Academy of Sciences\\
    \texttt{piotr.cezary.rybak@gmail.com}
}

\begin{document}
\maketitle
\begin{abstract}
Recently, open-domain question answering systems have begun to rely heavily on annotated datasets to train neural passage retrievers. However, manually annotating such datasets is both difficult and time-consuming, which limits their availability for less popular languages. In this work, we experiment with several methods for automatically collecting weakly labeled datasets and show how they affect the performance of the neural passage retrieval models. As a result of our work, we publish the MAUPQA dataset, consisting of nearly 400,000 question-passage pairs for Polish, as well as the HerBERT-QA neural retriever.
\end{abstract}

\section{Introduction}
\label{sec:intro}

Open-domain question answering (OpenQA) systems aim to provide answers to questions from a variety of topics, using a large collection of passages as a knowledge base. Recently, the development of such systems has been accelerated by the release of several large-scale question-passage datasets, such as MS MARCO \cite{nguyen2016ms}, TriviaQA \citep{joshi-etal-2017-triviaqa}, and Natural Questions \citep[NQ]{kwiatkowski-etal-2019-natural}. These datasets enabled the training of neural passage retrieval models (e.g. Dense Passage Retrieval, \citealp{karpukhin-etal-2020-dense}), which can select passages from a knowledge base that are the most likely to contain the answer to the question.

However, the annotation of such datasets is a time-consuming and expensive process, which limits their availability for less popular languages \citep{10.1145/3560260}. Another limiting factor is the availability of real questions. Datasets like MS MARCO or Natural Question consist of real questions asked by search engine users. For less popular languages (like Polish), such a source of questions is not available. This leads to two alternatives: either to train a system on a small dataset (which might not be sufficient for the model to reach its full potential) or to create a dataset automatically. The first approach was recently described by \citet{polqa} who published the PolQA dataset which consists of 7,000 trivia questions and 87,525 manually annotated passages.

In this work, we experiment with the latter approach and show how different methods for automatic data collection can impact the performance of the neural passage retrieval models. Our contributions can be summarized as follows:
\begin{enumerate}
    \item We experiment with several methods for automatically collecting weakly-labeled question-passage pairs, and show their impact on the performance of the retrieval models.
    \item We publish the MAUPQA dataset consisting of almost 400,000 question-passage pairs for Polish.\footnote{\url{https://hf.co/datasets/ipipan/maupqa}}
    \item We release the HerBERT-QA neural retriever, which achieves the best results on the PolQA dataset.\footnote{\url{https://hf.co/ipipan/herbert-base-qa-v1}}
\end{enumerate}

\begin{table*}[!ht]
\renewcommand*{\arraystretch}{1.2}
\setlength{\tabcolsep}{5pt}
\centering
\begin{tabular}{l|rrr|rrr|r}
    \toprule
    \bf{Dataset} & \bf{Questions} & \bf{Passages} & \bf{Answers} & \bf{Correct} & \bf{Unambiguous} & \bf{Relevant} & \bf{Overall} \\
    \midrule
    PolQA & 4,591 & 57,921 & 5,634 & 99\% & 99\% & 92\% & 90\% \\
    \midrule
    \midrule
    CzyWiesz-v2 & 29,078 & 29,078 & - & 100\% & 92\% & 73\% & 70\% \\
    GenGPT3 & 10,146 & 10,177 & 10,146 & 92\% & 44\% & 89\% & 33\% \\
    MKQA & 4,036 & 4,036 & 4,036 & 73\% & 73\% & 21\% & 15\% \\
    MTNQ & 135,781 & 142,008 & - & 60\% & 78\% & 80\% & 41\% \\
    MFAQ & 172,768 & 178,937 & - & 81\% & 84\% & 55\% & 43\% \\
    Templates & 15,993 & 15,993 & 14,520 & 88\% & 100\% & 89\% & 78\% \\
    WikiDef & 18,093 & 18,093 & 18,093 & 95\% & 77\% & 88\% & 65\% \\
    \midrule
    All & 385,895 & 398,322 & 46,795 & 76\% & 82\% & 69\% & 46\% \\
    \bottomrule
\end{tabular}
\caption{Basic statistics for all used datasets. \emph{All} represents the concatenation of all MAUPQA datasets (i.e. without PolQA). \emph{PolQA} refers to the training part of the PolQA dataset. \emph{PolQA} dataset has more answers than questions since it might contain multiple answer variants for a single question (e.g. \emph{7} and \emph{seven}). Some datasets don't have any answers due to the way they were created.} 
\label{tab:basic-stats}
\end{table*}

\section{Related Work}
\label{sec:related}

\paragraph{Weakly-labeled datasets} Over the years, many techniques were developed for the automatic creation of weakly-labeled datasets. One general idea is to use a weak model to automatically label the unlabeled dataset \citep{Lee2013PseudoLabelT}. In the case of OpenQA, either simple lexical models like BM-25 \citep{Robertson2009ThePR} or more powerful neural models are used to retrieve relevant passages for given questions. To further improve the accuracy of retrieved examples the passages can be filtered out using cross-encoders \citep{ren-etal-2021-rocketqav2} or answers (if available, \citealp{karpukhin-etal-2020-dense}).

However, the above method can only be used if the source of questions is available. If that is not the case, then questions can be automatically created. Either using templates \citep{fabbri-etal-2020-template} or trained models \citep{lewis-etal-2021-paq}.

Another line of work takes advantage of existing datasets and translates them automatically to other languages \citep{lewis-etal-2020-mlqa}. The quality of the machine translation model directly impacts the quality of the created dataset \citep{bonifacio2021mmarco}.

\paragraph{Polish OpenQA datasets} Few datasets exist for Polish OpenQA. The first published dataset for passage retrieval was the \emph{Czy wiesz?} dataset  \citep{marcinczuk-etal-2013-evaluation}. It is a collection of 4,721 questions from the \emph{Did you know?} section on Polish Wikipedia out of which only 250 questions were manually labeled with a relevant passage. \citet{rybak-etal-2020-klej} later annotated an additional 1,070 questions with relevant passages.

The PolQA dataset \citep{polqa} is a recently introduced dataset for Polish OpenQA. It consists of 7,000 trivia questions and 87,525 manually annotated passages (both positive and hard-negative). Even though the number of question-passage pairs is impressive for a less popular language like Polish, the number of unique questions is still rather limited.

\section{MAUPQA Dataset}
\label{sec:dataset}

The MAUPQA dataset consists of seven parts. Four of them are created from scratch (Czy wiesz?, GenGPT3, Templates, WikiDef), and the other three are based on existing resources (MKQA, MTNQ, MFAQ). 

\subsection{Quality Assessment}
To assess the quality of MAUPQA datasets, we sample and manually annotate 100 question-passage pairs for each dataset. Our manual verification consists of three aspects:

\paragraph{Correct} We check if the question is a valid, grammatically correct question written in Polish.

\paragraph{Unambiguous} We define that the question is ambiguous if it cannot be answered without providing additional information. For example, the question ``Where is the headquarter of the company?'' is ambiguous because it doesn't specify the name of the company and thus makes it impossible to answer the question.

\paragraph{Relevant} The final aspect is the relevance of the passage to the question, i.e. whether the passage contains the answer to the question.

\vspace{6pt}
We also calculate the \textbf{overall} correctness of the example as the proportion of examples that satisfy all three of the above aspects. We show the results of the quality evaluation in the Table \ref{tab:basic-stats} together with the sizes of each dataset.

\subsection{Datasets}
Below, we describe each of the seven MAUPQA datasets:

\paragraph{CzyWiesz-v2}
Similarly to the original \emph{Czy-wiesz?} dataset, we first gather all questions from the \emph{Did you know?} section on Polish Wikipedia together with a link to the relevant Wikipedia article. To select the relevant passage, we score all passages within this article using a multilingual cross-encoder \citep{bonifacio2021mmarco}\footnote{\url{https://hf.co/unicamp-dl/mMiniLM-L6-v2-mmarco-v2}} and choose the one with the highest score. We use a few simple heuristics to filter out questions regarding images (e.g. ``Who is the famous general \emph{in the photo}?''). Additionally, we remove questions from the KLEJ benchmark test set \citep{rybak-etal-2020-klej}.

The final dataset consists of 29,078 questions. They are grammatically correct, mostly unambiguous, and have a high rate of relevant passages (73\%, see Table \ref{tab:basic-stats}). Manual inspection shows that irrelevant passages are the result of the cross-encoder errors. In most cases, the relevant passage exists in the matching article but it was not selected.

\paragraph{GenGPT3}
In the GenGPT3 dataset, we explore the application of the \emph{text-davinci-003} model \citep{ouyangtraining} for generating question-answer pairs based on a given passage. To obtain passages, we use the Polish subset of CCNet \citep{wenzek-etal-2020-ccnet}. These passages turned out to be very diverse, covering domains such as news, legal, technical, etc. To guide the model in generating relevant questions, we use the prompt: \emph{Napisz pytanie i odpowiedź do poniższego paragrafu. Pytanie musi mieć przynajmniej pięć słów. Odpowiedź może mieć najwyżej pięć słów} (Write a question and answer for the following passage. The question must be at least five words. The answer can be up to five words). In addition, we provide two examples within the prompt to help the model learn to generate appropriate question-answer pairs.

Through our experiments, we observe that the generated questions are grammatically correct in 92\% of the cases and highly relevant (89\% of the cases). However, we also find that the questions are often ambiguous, with 56\% of them requiring a contextual understanding of the passage to answer.

\paragraph{MKQA}
The MKQA \citep{longpre-etal-2021-mkqa} dataset consists of 10,000 questions sampled from the NQ dataset and manually translated into 25 languages (including Polish). We clean MKQA dataset by removing questions without answers, requiring long answers (\emph{Why?} and \emph{How?} questions), and ambiguous ones (``Who is the \emph{current} president?''). We end up with 4,036 questions.

Since the original dataset doesn't include matching passages, we use the BM-25 algorithm \citep{Robertson2009ThePR} to select the top 100 candidate passages which we re-rank using a multilingual cross-encoder. In either case, we append the answer to the query to increase the performance of the passage retrieval. However, it still proved to be difficult to retrieve relevant passages and only 21\% of them are correct.

\paragraph{MTNQ}
To create the machine-translated NQ dataset (MTNQ) we select all questions with relevant passages from the NQ dataset and split those passages into sentences. Then, we translate both questions and sentences into Polish using Allegro\footnote{\url{https://ml.allegro.tech/}} machine translation model.

Even though the translation model is high quality (similar to Google Translate), the translations still contain many errors. Two main reasons are incorrectly translating named entities (e.g. movie titles) and very noisy input (NQ questions are Google search phrases). It is worth noting that MKQA, which is a manually translated subset of NQ, also has a high ratio of ungrammatical questions.

\paragraph{MFAQ}
The MFAQ dataset \citep{de-bruyn-etal-2021-mfaq} contains 234 million multilingual (4 million Polish) questions scraped from FAQ websites. However, many of them are artificially created, e.g. ``What is the best hotel in \emph{city}?'' for hundreds of different \emph{cities}. To clean the data, we cluster lexically similar questions and passages and remove clusters with over 5 questions. Additionally, some of the questions are not in Polish. We filter them using the fasttext language-id model \citep{joulin-etal-2017-bag,joulin2016fasttext}. 

After filtering, the dataset contains 178,937 passages, i.e. less than 5\% of the original dataset. This shows the risk of using questions extracted directly from crawled websites. The cleaned dataset has rather high quality, in terms of grammatical correctness, unambiguity, and relevance of passages. The MFAQ is much more diverse than other datasets (except for \emph{GenGPT3}) and contains questions from a wide range of domains (customer support, lifestyle, technical, etc.).

\paragraph{Templates}
We take advantage of the Wikipedia structure to generate questions using predefined templates. For example, list pages group together similar entities (e.g. ``Writers born in Poland'') which allows generating questions like ``Where was Zbigniew Herbert born?''. We also use tables (e.g. ``What is the capital of Poland?'') and chronologies (e.g. ``In which year World War 2 started?''). In total, we use 33 templates to generate questions. Since each question has a link to the relevant Wikipedia article, we use the same method as in the \emph{CzyWiesz-v2} dataset to select the most relevant passage from the relevant article.

Overall, we created 15,993 questions from templates. They are high quality but the process of creating templates was surprisingly time-consuming and took a few hours per template.

\paragraph{WikiDef}
We use Wiktionary\footnote{\url{https://www.wiktionary.org/}} to generate questions based on word definitions. Some definitions have links to Wikipedia articles which we use to create the question-passage pairs. For example, the definition of ``Monday'' is ``the first day of the week''. Based on it, we generate the question ``What is the name of \emph{the first day of the week}?''. Then, we select the first passage from the linked Wikipedia article as the relevant passage. We remove short definitions (less than 5 words) containing names of 23 predefined ``categories`` (e.g. city) to avoid ambiguous questions (e.g. ``What is the name of \emph{a city in Poland}?'').

We end up with 18,093 questions asking for word definitions. This is the least diverse dataset of all as all questions follow the same template. Even though we tried to filter unambiguous questions there are still 23\% of them in the final dataset.

\section{Evaluation}
\label{sec:eval}
We use the Tevatron library \citep{Gao2022TevatronAE} to train the neural retriever. For each dataset, we fine-tune the HerBERT Base model \citep{mroczkowski-etal-2021-herbert} for 2,000 steps, with batch size 128 and learning rate $10^{-5}$. Otherwise, we use default parameters. We experimented with training models for 5,000 steps but it didn't increase the performance. We use a single hard-negative per question when training on PolQA dataset. For other datasets, we only use in-batch random negatives as they don't contain hard-negatives.

For evaluation, we use Accuracy@10 (i.e. is there at least one relevant passage within the top 10 retrieved passages) and NDCG@10 (i.e. score of each relevant passage within the top 10 retrieved passages depends descending on its position, \citet{ndcg}). Each model is evaluated on the PolQA development dataset. We use provided Polish Wikipedia dump as a knowledge base.

\section{Results}
\label{sec:results}
The baseline retriever trained using manually annotated PolQA dataset achieves 60.8\% accuracy@10 (see Table \ref{tab:results}). Individually, none of the automatically created datasets has a comparable score. 

As expected, the best model is \emph{MTNQ} with an accuracy of 58.5\%. It is the second largest dataset, similarly to PolQA it contains mostly trivia questions, and is based on manually labeled question-passage pairs. Comparably large \emph{MFAQ} dataset obtains much lower performance (38.7\%), probably due to domain mismatch as otherwise, its quality is higher than \emph{MTNQ}.

The \emph{MKQA}, which is a manually translated subset of \emph{NQ} dataset achieves surprisingly good results (51.5\%). It is unexpected considering that only 21\% of its passages are actually relevant.

The second best result (54.2\%) is achieved by the \emph{GenGPT3} dataset. Despite the diverse nature of the questions from different domains, and the relatively modest size of the dataset, it exhibits a remarkable level of quality that allows it to serve as a reliable source for training a passage retriever.

The third best result (54.1\%) is scored by \emph{CzyWiesz-v2} dataset. The other two datasets created based on Wikipedia perform much worse, \emph{Templates} obtains accuracy of 45.9\% and \emph{WikiDef} only 19.9\%. It is also the lowest result of all datasets, probably due to its low diversity.

None of the datasets is perfect and each of them has its own disadvantages. However, the retriever trained on all of them results in better performance than the manually annotated dataset (61.2\% vs 60.8\%). If we further fine-tune the retriever pre-trained on MAUPQA, we obtain the state-of-the-art result for Polish passage retrieval of 62.7\%. We name this retriever HerBERT-QA and release it alongside the created datasets.

\begin{table}[!ht]
\renewcommand*{\arraystretch}{1.2}
\setlength{\tabcolsep}{10pt}
\centering
\begin{tabular}{l|rrr|r}
    \toprule
    \bf{Dataset} & \bf{Acc@10} & \bf{NDCG@10} \\
    \midrule
    PolQA & 60.8\% & 26.9\% \\
    \midrule
    \midrule
    CzyWiesz-v2 & 54.1\% & 22.0\% \\
    GenGPT3 & 54.2\% & 22.1\% \\
    MKQA & 51.5\% & 21.6\% \\
    MTNQ & 58.5\% & 24.1\% \\
    MFAQ & 38.7\% & 14.0\% \\
    Templates & 45.9\% & 16.9\% \\
    WikiDef & 19.9\% & 7.7\% \\
    \midrule
    All & 61.2\% & 25.2\% \\
    All $\rightarrow$ PolQA & \bf{62.7\%} & \bf{27.4\%} \\
    \bottomrule
\end{tabular}
\caption{Passage retriever performance trained on different datasets. We use top-10 accuracy and NDCG@10 on the PolQA development set. \emph{All} represents the concatenation of all MAUPQA datasets (i.e. without PolQA). \emph{All $\rightarrow$ PolQA} is a model first trained on the MAUPQA dataset and then fine-tuned on the PolQA dataset.}
\label{tab:results}
\end{table}

\section{Conclusion}
\label{sec:conclusion}
In this work, we present MAUPQA, the largest Polish QA dataset with almost 400k question-passage pairs. Even though the dataset is created automatically it achieves competitive results on the Polish passage retrieval task and after fine-tuning on the PolQA dataset sets a new state-of-the-art performance.

Each of the seven datasets which make up MAUPQA has different properties and results in the vastly different performance of passage retrievers. Thanks to recent advancements of machine translation models, we recommend translating existing English datasets as the best way to cheaply obtain competitive QA datasets. Otherwise, generating questions using GPT-3 model proves to work well and can be applied to multiple different domains (for which there might not be an English dataset). If a set of questions already exists for a given language, then using pseudo-labeling also results in a surprisingly good dataset. However, to get the best performance, it is useful to combine multiple different datasets.

We believe our work will benefit the Polish NLP community, both by publishing a MAUPQA dataset, as well as the state-of-the-art passage retrieval model. Our study also lays a path for other languages on how to construct similar datasets.

\section*{Limitations}
The MAUPQA dataset focuses only on the Polish language and the drawn conclusions might not hold for other languages. For example, the format of sentences in the \emph{Did you know?} section of Polish Wikipedia makes it very easy to transform them into questions. This is not the case for other languages. Some of them don't even have the \emph{Did you know?} section.

Except for choosing the number of training steps (2,000 or 5,000), we didn't perform any additional hyper-parameter search and used the default Tevatron values. We also tested only one encoder architecture (HerBERT Base). The results for other setups might be different.

Except for GenGPT3 and MFAQ, all datasets (including the evaluation dataset) use Wikipedia as a knowledge base. This might negatively impact the perceived performance of the retrievers trained on GenGPT3 and MFAQ. We suspect that those retrievers might generalize better to other domains but there are no Polish QA datasets on which we could have tested it.

\section{Acknowledgments}
We thank the Allegro.com Machine Learning Research team for giving us access to their machine translating model.

This work was supported by the European Regional Development Fund as a part of 2014–2020 Smart Growth Operational Programme, CLARIN — Common Language Resources and Technology Infrastructure, project no. POIR.04.02.00-00C002/19.

\bibliography{anthology,custom}
\bibliographystyle{acl_natbib}

\end{document}